\documentclass[sigconf]{acmart}
\AtBeginDocument{%
  }
    
\copyrightyear{2025}
\acmYear{2025}
\setcopyright{cc}
\setcctype{by-nc}
\acmConference[WWW '25]{Proceedings of the ACM Web Conference 2025}{April 28-May 2, 2025}{Sydney, NSW, Australia}
\acmBooktitle{Proceedings of the ACM Web Conference 2025 (WWW '25), April 28-May 2, 2025, Sydney, NSW, Australia}
\acmDOI{10.1145/3696410.3714842}
\acmISBN{979-8-4007-1274-6/25/04}

\usepackage{bbding}
\usepackage{algorithm}
\usepackage{algorithmic}
\usepackage{subfigure}
\usepackage{multirow}
\usepackage{balance}

\usepackage{fancybox}

\newcommand{\FancyBox}[2]{%
    \noindent
    \fbox{
        \begin{minipage}{0.96\linewidth}
            \vspace{2pt}
            \textbf{#1} 
            \vspace{2pt}
        \end{minipage}
    }
    \par
    \noindent
    \doublebox{
        \begin{minipage}{0.96\linewidth}
            \vspace{5pt}
            #2 
            \vspace{5pt}
        \end{minipage}
    }
}

\newcommand{\eg}{\emph{e.g., }}

\begin{document}

\title{Large Language Models Empowered Personalized Web Agents}

\author{Hongru Cai}
\email{henry.hongrucai@gmail.com}
\affiliation{
\institution{National University of Singapore}
\country{Singapore}}

\author{Yongqi Li}
\authornote{Corresponding authors.}
\email{liyongqi0@gmail.com}
\affiliation{
\institution{The Hong Kong Polytechnic University}
\city{Hong Kong SAR}
\country{China}}

\author{Wenjie Wang}
\email{wenjiewang96@gmail.com}
\authornotemark[1]
\affiliation{
\institution{University of Science and Technology of China}
\city{Hefei}
\country{China}}

\author{Fengbin Zhu}
\email{zhfengbin@gmail.com}
\affiliation{
\institution{National University of Singapore}
\country{Singapore}}

\author{Xiaoyu Shen}
\email{xyshen@eitech.edu.cn}
\affiliation{
\institution{Eastern Institute of Technology, Ningbo}
\city{Ningbo}
\country{China}}

\author{Wenjie Li}
\email{cswjli@comp.polyu.edu.hk}
\affiliation{
\institution{The Hong Kong Polytechnic University}
\city{Hong Kong SAR}
\country{China}}

\author{Tat-Seng Chua}
\email{dcscts@nus.edu.sg}
\affiliation{
\institution{National University of Singapore}
\country{Singapore}}
\renewcommand{\shortauthors}{Hongru Cai et al.}

\begin{abstract}
Web agents have emerged as a promising direction to automate Web task completion based on user instructions, significantly enhancing user experience. 
Recently, Web agents have evolved from traditional agents to Large Language Models (LLMs)-based Web agents. 
Despite their success, existing LLM-based Web agents overlook the importance of personalized data (\eg user profiles and historical Web behaviors) in assisting the understanding of users' personalized instructions and executing customized actions. 

To overcome the limitation, we first formulate the task of LLM-empowered personalized Web agents, which integrate personalized data and user instructions to personalize instruction comprehension and action execution. 
To address the absence of a comprehensive evaluation benchmark, we construct a \textbf{Personal}ized \textbf{W}eb \textbf{A}gent \textbf{B}enchmark (PersonalWAB), featuring user instructions, personalized user data, Web functions, and two evaluation paradigms across three personalized Web tasks. 
Moreover, we propose a \textbf{P}ersonalized \textbf{U}ser \textbf{M}emory-enhanced \textbf{A}lignment (PUMA) framework to adapt LLMs to the personalized Web agent task. PUMA utilizes a memory bank with a task-specific retrieval strategy to filter relevant historical Web behaviors. 
Based on the behaviors, PUMA then aligns LLMs for personalized action execution through fine-tuning and direct preference optimization. 
Extensive experiments validate the superiority of PUMA over existing Web agents on PersonalWAB. 
We release code and data at \href{https://hongrucai.github.io/PersonalWAB/}{PersonalWAB github repository}. 
\end{abstract}

\ccsdesc[500]{Information systems~Web applications}
\ccsdesc[500]{Information systems~Personalization}

\keywords{Personalized Web Agents, Personalization, Large Language Model}


\maketitle

\section{Introduction}

\begin{figure}[t]
\setlength{\abovecaptionskip}{-0.10cm}
\setlength{\belowcaptionskip}{-0.20cm}
\centering
\includegraphics[width=1\linewidth]{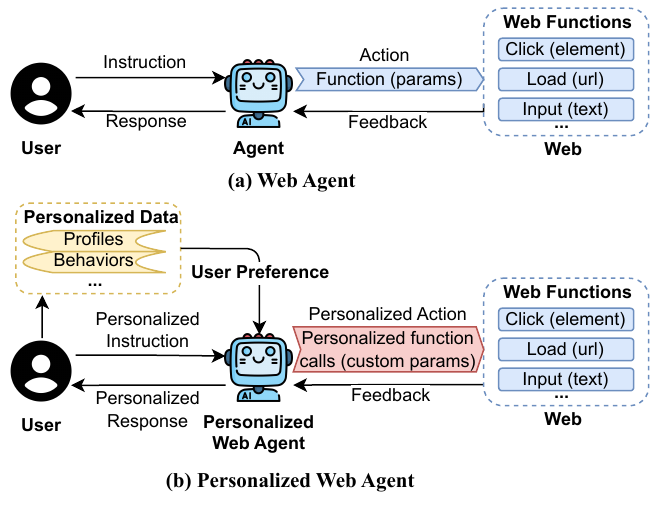}
\caption{Comparison between traditional Web agents (a) and personalized Web agents (b). Personalized Web agents adopt personalized data to infer implicit user preferences, assisting in understanding user instructions and performing personalized actions, leading to more intelligent Web services.}
\vspace{-1.3em}
\label{Fig: intro}
\end{figure}

The World Wide Web has evolved into a fundamental infrastructure in the information age, with diverse Web services integrated into users' daily lives, including information retrieval, online shopping, and social engagement. 
However, the unprecedented scale and complexity of modern Web services also present new challenges. Users, particularly the elderly groups, are overwhelmed with vast amounts of unstructured data and intricate interactions, complicating task completion on the Web~\cite{2021russ}. 
To alleviate the burden of complex Web operations, Web agents have emerged as a promising solution~\cite{2017worldofbits} to bridge users and Web services as shown in Figure~\ref{Fig: intro}(a). Based on user instructions, Web agents autonomously interact with the Web to complete tasks such as information retrieval and online shopping~\cite{2022webshop}, offering a convenient way to enhance efficiency and intelligence with extensive Web services. 

The evolution of Web agents has undergone a significant transition from traditional agents to those powered by Large Language Models (LLMs). Traditional agents typically optimize Web navigation tasks by reinforcement learning techniques~\cite{2017worldofbits, 2018miniwob, 2022webshop}, while their context understanding and reasoning capabilities are limited, failing to generalize to complex and out-of-distribution scenarios~\cite{deng2023mind2web}. 
In recent years, LLMs have demonstrated extensive world knowledge along with strong understanding, planning, and reasoning capabilities, making LLM-based Web agents a rapidly evolving direction~\cite{openai_gpt-4_2024, dubey2024llama3}. 
Related research has leveraged techniques such as in-context learning~\cite{2024largellmcomputer, 2024adaplanner, zheng2024synapse,zheng2024seeact}, fine-tuning~\cite{deng2023mind2web, gur2024realwebagent}, and reinforcement learning~\cite{putta2024agentq} to enhance the instruction-following capabilities of LLMs in various Web agent tasks. 
Notably, in addition to single-turn user instructions, some studies have explored utilizing the powerful interactive capabilities of LLMs to enable multi-turn interactions with users, facilitating conversational Web navigation and task execution~\cite{sun-etal-2022-metagui,lù2024weblinx, deng-2024-multiturnweb}.

Despite the significant success of LLM-based Web agents, they overlook the critical role of personalized data in enhancing user experience, as illustrated in Figure~\ref{Fig: intro}. 
Personalized data, such as user profiles and historical Web behaviors, reveals implicit user preference, which can facilitate the understanding of user instructions and enable personalized action execution. 
Specifically, 1) personalized data can supplement user context for \textit{\textbf{personalized instruction comprehension}}. For example, when users search for a product, their behavior history may reveal implicit preferences for product attributes (\eg price) that are not explicitly stated in user instructions. Besides, 2) personalized data enables \textit{\textbf{personalized action execution}}, where actions can be formulated as various Web function calls from Web services~\cite{2022webshop,zhou2024webarena,koh2024visualwebarena,zhang2024mmina}. 
In this work, we abstract the Web services from various websites (\eg, Amazon shopping website) as a diverse set of \textit{``Web functions\footnote{``Web APIs'' and ``Web tools'' are also used to convey similar meanings in agents. For convenience, we unify these terms as ``Web functions'' below.}''}; for example, the ``search'' action can be executed by passing a textual query as function parameters to the search API of a website. 
Users have varying habits and preferences for Web services, leading to personalized function calls with customized parameters.

In this light, we formulate the task of \textbf{LLM-empowered personalized Web agents}, which integrate personalized user data for personalized instruction comprehension and action execution, aiming to align with explicit user instructions and implicit user preferences derived from personalized data. 
Formally, given user instructions alongside personalized user data (\eg profiles and historical Web behaviors), LLMs must infer personalized user requirements and preferences to determine which Web function to call and formulate the corresponding function parameters. 
Subsequently, the results of these function calls are returned to users. 
However, to advance this task, the primary challenge is the lack of a comprehensive benchmark for training and evaluation.

To bridge this gap, we construct the first \textbf{Personal}ized \textbf{W}eb \textbf{A}gent \textbf{B}enchmark (PersonalWAB). PersonalWAB focuses on three representative tasks for personalized Web agents: personalized search, recommendation, and review generation on Web platforms, which require LLMs to infer user preferences for task completion. Specifically, PersonalWAB is constructed by the following steps: 
\setlength{\leftmargini}{10pt}
\begin{itemize}
\vspace{-0.1cm}

    \item [1)] \textit{Personalized Data Construction}: PersonalWAB adopts Amazon review dataset~\cite{hou2024bridging} to construct 1,000 diverse users with simulated profiles and real Web behaviors (\eg purchase and rating).

    \item [2)] \textit{User Instruction Creation}: For the three tasks, PersonalWAB utilizes users' genuinely liked items as the ground truth for search and recommendation, and real reviews as the ground truth for review generation. PersonalWAB then uses ground truth items and reviews to synthesize corresponding user instructions. 
    
    \item [3)] \textit{Web Environment Implementation}: To interact with the Web environment, PersonalWAB develops a series of Web functions for the three tasks. 
    
    \item [4)] \textit{Evaluation}: PersonalWAB utilizes the ground truth in Step 2 to assess the three tasks. Notably, it not only supports single-turn evaluation but also develops an LLM-based user simulator for real-time multi-turn interaction and evaluation with users. 
    
\vspace{-0.12cm}
\end{itemize}
Extensive analysis in Section~\ref{sec:benchmark_analysis} validates that PersonalWAB offers a set of users with diverse profiles and behaviors, and the simulated user profiles closely align with the actual behaviors by empirical evaluation. 
The comparison with existing benchmarks is in Table~\ref{tab: Benchmark_compare}. 

\begin{table}[]
\centering
\caption{Comparison between existing benchmarks and PersonalWAB from three key aspects: interaction type with users, Web environment type for agent interactions, and utilization of personalized data.}
\vspace{-1em}
\setlength{\tabcolsep}{1mm}{
\resizebox{0.48\textwidth}{!}{
\begin{tabular}{cccc}
\hline
\multicolumn{1}{c}{Benchmark} &
  \multicolumn{1}{c}{\begin{tabular}[c]{@{}c@{}}Interaction\\ Type\end{tabular}} &
  \multicolumn{1}{c}{\begin{tabular}[c]{@{}c@{}}Environment\\ Type\end{tabular}} &
  \multicolumn{1}{c}{Personalization} \\ \hline
MiniWoB++~\cite{2018miniwob}   & Single-turn                & Mobile Web UI   & \textcolor{red}{$\times$}               \\
RUSS~\cite{2021russ}        & Multi-turn               & Web UI          & \textcolor{red}{$\times$}               \\
META-GUI~\cite{sun-etal-2022-metagui}    &  Multi-turn               & Mobile apps                & \textcolor{red}{$\times$}      \\
WebShop~\cite{2022webshop}     & Single-turn               & Shopping Web UI & \textcolor{red}{$\times$}               \\
Mind2Web~\cite{deng2023mind2web}    & Single-turn                & Web UI          & \textcolor{red}{$\times$}               \\
WebArena~\cite{zhou2024webarena}    & Single-turn                & Web UI           & \textcolor{red}{$\times$}               \\
VWA~\cite{koh2024visualwebarena}   & Single-turn                & Web UI           & \textcolor{red}{$\times$}               \\
WebVoyager~\cite{he2024webvoyager}  & Single-turn                & Web UI           & \textcolor{red}{$\times$}               \\
WorkArena~\cite{workarena2024}   &  Multi-turn               & Web UI           & \textcolor{red}{$\times$}               \\
WebLINX~\cite{lù2024weblinx}     &  Multi-turn               & Web UI           & \textcolor{red}{$\times$}               \\
MT-Mind2Web~\cite{deng-2024-multiturnweb} &  Multi-turn               & Web UI           & \textcolor{red}{$\times$}               \\
MMInA~\cite{zhang2024mmina}       & Single-turn                & Web UI          & \textcolor{red}{$\times$}               \\
Turking~\cite{xu2024turkingbench}     & Single-turn                & Web UI           & \textcolor{red}{$\times$}               \\
ChatShop~\cite{chen2024chatshop}    &  Multi-turn               & Web function         & \textcolor{red}{$\times$}               \\ \hline
PersonalWAB & \multicolumn{1}{c}{\begin{tabular}[c]{@{}c@{}}Single-turn \\ \& Multi-turn\end{tabular}}               & Web function         & \textcolor{green}{\Checkmark}               \\ \hline
\end{tabular}
}}
\vspace{-1.5em}
\label{tab: Benchmark_compare}
\end{table}

To enable LLM-empowered personalized Web agents, we propose a \textbf{P}ersonalized \textbf{U}ser \textbf{M}emory-enhanced \textbf{A}lignment (PUMA) framework. 
PUMA stores users' long-term Web behaviors into a memory bank and utilizes a task-specific retrieval strategy to filter out irrelevant information, focusing only on behaviors and features relevant to the current instruction. 
Given the retrieved behaviors and features, PUMA then combines them with user instructions to call appropriate Web functions and generate optimal function parameters to enhance the returned results. 
However, the large parameter space challenges LLMs in producing high-reward parameters. 
To address this, PUMA designs several heuristic strategies to construct pseudo-label parameters for supervised fine-tuning (SFT)~\cite{radford2018improving}, enabling 
LLMs to generate reasonable function parameters. 
PUMA then uses Direct Preference Optimization (DPO)~\cite{2023dpo} to sample multiple function parameters for pair-wise optimization, better aligning with personalized user preferences. 
Experimental results demonstrate that PUMA significantly outperforms existing Web agents in single-turn and multi-turn personalized Web tasks, showcasing the potential of personalized Web agents to deliver more intelligent, customized, and user-centered Web services. 

The key contributions in this work are as follows:
\vspace{-0.5em}
\setlength{\leftmargini}{10pt} \begin{itemize}
\item We are the first to formulate the task of LLM-empowered personalized Web agents, which incorporate personalized user data to achieve personalized instruction understanding and action execution, bridging users with customized Web services. 

\item We construct the first benchmark for LLM-empowered personalized Web agents, featuring a diverse set of users with varying profiles and behaviors, the instructions across three tasks, callable Web functions, and two evaluation paradigms. 

\item We propose PUMA, a novel personalized alignment framework with a user memory bank and optimization strategies to align LLMs with the task of personalized Web agents.

\item We conduct extensive experiments on PersonalWAB, showing that PUMA consistently surpasses existing Web agents, aligning better with personalized user instructions and preferences.
\vspace{-1em}
\end{itemize}
\section{Related Work}
\noindent$\bullet$ \textbf{Web agents.} Web agents are designed to automate a variety of complex Web-based tasks. 
Some studies focus on directly responding to users' instructions in a single turn.
MiniWoB++~\cite{2018miniwob} established a platform of website widgets where agents can complete online tasks through keyboard and mouse.
Webshop~\cite{2022webshop} introduced a simulated e-commerce environment with human-written task instructions.
Recent studies investigate automating Web tasks under more practical and complex settings, including multi-domain~\cite{deng2023mind2web}, multi-hop~\cite{zhang2024mmina}, real-time interactions with Web~\cite{zhou2024webarena}, and visual UI understanding~\cite{koh2024visualwebarena, he2024webvoyager}.
Numerous efforts have been made to solve these problems, including fine-tuning~\cite{nakano2021webgpt, gur2024realwebagent, furuta2024webgum} and prompting LLMs~\cite{yao2023react, 2024reflexion, zheng2024seeact}.
2) Another research direction involves integrating user interactions into the agent's execution process.
META-GUI~\cite{sun-etal-2022-metagui} introduced a dataset focused on automating actions in mobile apps following conversational user instructions.
RUSS~\cite{2021russ} designed a dataset to boost dialogue-centric Web navigation.
Recent works also focus on conversational Web navigation~\cite{lù2024weblinx, deng-2024-multiturnweb} and interactive information-seeking problems~\cite{chen2024chatshop}.

Despite advancements, prior research overlooks the dimension of personalization in Web agents. 
A recent study simulates users with distinct roles, permissions, and interaction histories~\cite{zhou2024webarena}, but these roles are predefined per platform and do not require understanding user preferences nor demand the agent to adjust the execution strategy according to user preferences.
In this work, we first focus on LLM-empowered personalized Web agents and propose a novel framework along with a benchmark for the optimization and evaluation of LLM-empowered personalized Web agents. 

\noindent$\bullet$ \textbf{Personalized LLMs.}
Personalized LLMs are designed to handle user personas (\eg, background information or historical behaviors) to meet individualized needs, adapting to distinct users~\cite{tseng2024personallmssurvey}.
Research in this field falls into two main categories: personalized content generation and user-facing applications. 1) Personalized content generation focuses on the core challenges of generating personalized content. 
They have used openly available user data on Reddit~\cite{welch-etal-2022reddit}, Facebook, Twitter~\cite{2017twitter}, and other blogging websites~\cite{king-cook-2020blog} to pre-train LLMs.
Key tasks include stance classification, demographic inference~\cite{soni-etal-2022-human}, and personalized sentiment prediction~\cite{mireshghallah-etal-2022-useridentifier}.
Benchmarks like LaMP~\cite{salemi2024-lamp} and LongLaMP~\cite{kumar2024longlamp} further provide datasets for evaluating personalized text classification and content generation. 
2) Another research direction is the practical applications in real-world scenarios, starting with personalized dialogue systems. 
Studies have built dialogue datasets by promoting crowd-workers to author dialogues based on specific personas~\cite{zhang-etal-2018-personalizing}, and by extracting user attributes from Reddit~\cite{mazare-etal-2018-training} and Weibo~\cite{zhong-etal-2022-less}.
Apollonion~\cite{chen2024apollonion} dynamically updates user profiles for personalized responses.
Additionally, memory mechanisms~\cite{lu2023memochattuningllmsuse,lee-etal-2023-prompted, wu2023autogenenablingnextgenllm} help models recall past conversations and important events. 
Personalized LLMs are also applied in healthcare~\cite{abbasian2024opencha, jin2024healthllm}, education~\cite{2023educhat,shehata-etal-2023-HumSum}, and robotics~\cite{2023tidybot} to enhance services.

However, previous studies have not explored personalized function calls tailored to user-specific needs. 
Our work bridges this gap by emphasizing adapting agents' actions to different users by utilizing personalized user data and enabling a comprehensive assessment of agents’ ability to complete several personalized tasks in Web environments.  
\section{Task and Benchmark}
In this section, we formulate the task of LLM-empowered Web agents in Section~\ref{task fomulation}, detail the construction of PersonalWAB in Section~\ref{sec:personalWAB}, and present the evaluation paradigms in Section~\ref{sec:evaluation}. 
\begin{figure*}[]
\setlength{\abovecaptionskip}{0.1cm}
\setlength{\belowcaptionskip}{0cm}
\centering
\includegraphics[width=1\linewidth]{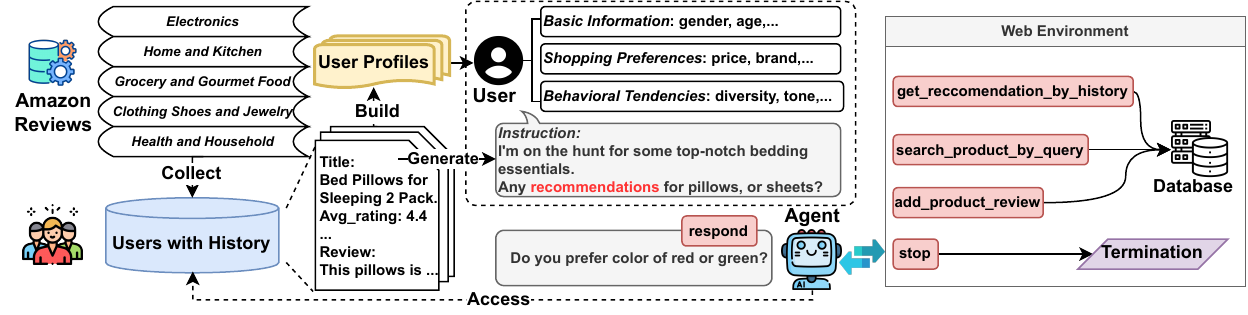}
\caption{Overall pipeline for constructing PersonalWAB benchmark with a real example for recommendation instruction.}
\vspace{-1em}
\label{fig: benchmark_construction}
\end{figure*}

\subsection{Task Formulation}
\label{task fomulation}

LLM-empowered personalized Web agents act as intermediaries between users and Web services, and we formulate the following elements in this task:
\setlength{\leftmargini}{10pt} \begin{itemize}
\vspace{-0.5em}
\item User. Each user $ u \in \mathcal{U} $ has a distinct profile $ P_{u} $ and the historical Web behaviors $ H_{u} $. 
The profile $ P_{u} $ includes static attributes such as demographics, and $ H_{u} $ records the user's past Web behaviors, represented as a time-ordered sequence, $ \{ h_{u}^1, h_{u}^2, \dots, h_{u}^N \} $. Each  $ h_{u}^i $ denotes one Web behavior, such as a purchase or a review.
\item Instruction. The user's instruction \( i_{u} \) is a natural language sentence that expresses their needs and requirements.
\item Web environment. It is abstracted as a series of Web functions, denoted by  $ \mathcal{T} $. Each function $ f \in \mathcal{T} $ can be invoked with an input parameter $ p $, returning the corresponding result $ O_{f_p} $. Notably, different input parameters will yield different function results.  
\end{itemize}
\vspace{-1em}
Given the user instruction \( i_{u} \) and the personalized data $ P_{u} $ and  $ H_{u} $, LLM-empowered personalized Web agents aim to select the appreciate Web function $ f $ and determine the optimal parameter $ p $ to invoke personalized results $ O_{f_p} $ from the Web environment.

\subsection{Benchmark Construction}\label{sec:personalWAB}

It is challenging to gather a set of users to collect the real data on Web agent applications. 
Therefore, we chose to develop the benchmark using existing datasets of users' Web behaviors. 
Specifically, we selected the Amazon Review~\cite{hou2024bridging} dataset as the foundation for our benchmark, as it provides a large-scale collection of users' Web behaviors, including purchase and product rating across various categories of products. The overall pipeline for constructing PersonalWAB is illustrated in Figure~\ref{fig: benchmark_construction}, which consists of three steps: \textit{personalized data construction}, \textit{user instruction creation}, and \textit{Web environment implementation}.

\subsubsection{Personalized Data Construction}
This section consists of user sampling and user profile generation steps. 

\vspace{3pt}
\noindent$\bullet$ \textbf{User sampling.} We randomly selected 1,000 users from the Amazon Review across five distinct product categories: \textit{Electronics}, \textit{Home and Kitchen}, \textit{Grocery and Gourmet Food}, \textit{Clothing, Shoes, and Jewelry}, and \textit{Health and Household}.
For each user, we collected all their interactions across the five categories, each containing detailed purchased product information (such as product title, price, rating, and store) and user evaluations (including ratings, titles, and comments).

To simulate realistic user behavior, we first arranged all user interactions chronologically and divided them based on timestamps into three segments: 80\% as historical data, 10\% for the training set, and the final 10\% for the test set.

\vspace{3pt}
\noindent$\bullet$ \textbf{User profile generation.} 
We generated unique profiles for each of the 1,000 users based on their entire history of behaviors, using the language model to infer and summarize their potential profiles (see prompt in Figure~\ref{app: prompt profile generation}).
Each user profile is structured to reflect the following key dimensions (see details in Figure\ref{app: prompt profile generation cho}):

\setlength{\leftmargini}{10pt} \begin{itemize}

\item Basic information.
This includes fundamental personal attributes such as \textit{gender}, \textit{age}, and \textit{occupation}, inferred from the user's product categories and purchasing behaviors.

\item Shopping preferences.
This dimension captures the user's purchasing tendencies, including 1) \textit{price sensitivity} (whether the user gravitates towards budget, mid-range, or premium products), 2) \textit{shopping interests} (the types of products the user is most frequently drawn to), and 3) \textit{brand preferences} (the brands most commonly referenced in the user’s purchase history). 

\item Behavioral tendencies.
We summarize the characteristics of each user's shopping behavior using LLM from the following aspects.
1) \textit{Diversity preference} indicates whether the user favors trying new products or sticking with familiar ones;
2) \textit{Interaction complexity} describes whether the user prefers concise or detailed interactions based on their review patterns; 
3)\textit{Tone and style} capture the emotional tone and expressive style of the user's reviews, which may affect how they engage with systems. 
4) \textit{Item reference} examines how often the user refers to specific products or brands in their reviews; and 5) \textit{focus aspects} highlight which product features (e.g., price, average rating, brand) the user tends to prioritize in their reviews.

\end{itemize}

The user profiles will support the following personalized instruction generation (\S~\ref{pertaskgen}) and multi-turn evaluation (\S~\ref{multiturn eval}).

\subsubsection{User Instruction Creation}
\label{pertaskgen}

As previously mentioned, organizing thousands of users to collect their real instructions poses significant challenges. 
To address this, we prompt the LLM to generate personalized instructions based on each user's profile and real Web behaviors. These instructions encompass three task scenarios: search, recommendation, and review.

\setlength{\leftmargini}{10pt} \begin{itemize}

\item Search instructions: 
We provide the language model with a detailed user profile and product information, including key attributes like brand, category, and features, to generate instructions for searching products. 
The prompt is detailed in Figure~\ref{app: prompt search generation}. 
Depending on the profile, the generated search instructions vary in length, tone, level of detail regarding the product, and the specific product aspects mentioned.

\item Recommendation instructions: 
The recommendation instruction requests generated tend to be shorter, more general, and less specific, leaving room for broader exploration. We prompt the LLM to generate recommendation tasks with the prompt (see Figure~\ref{app: prompt rec generation}), user profile, and the user's integrated products.

\item Review instructions: 
The LLM receives both the user profile, target product information, and actual review text to generate user instructions for writing the review with users' personalized requirements. 
The prompt details are shown in Figure~\ref{app: prompt review generation}. 

\end{itemize}

\subsubsection{Web Environment Implementation}
We choose to abstract and simplify the Web environment as a series of Web functions~\cite{chen2024chatshop} rather than Web GUIs~\cite{xu2024turkingbench}, as we believe GUIs are primarily user-friendly interfaces for humans and not essential for agents. 
The following web functions have been developed to help the agent complete users' instructions:

\setlength{\leftmargini}{10pt} \begin{itemize}
\item \textbf{search\_product\_by\_query.} 
This function takes a textual query as input and returns detailed information on the 10 most similar products based on the query. 
We facilitate this function using BM25 with Pyserini~\cite{2021pyserini} to enable fast retrieval from a database of all products.
\item \textbf{get\_recommendations\_by\_history.} 
This function accepts a sequence of product IDs and returns 10 recommended products.
To implement this, we trained the SASRec model~\cite{2018sasrec} on our conducted benchmark, with cold-start products removed.
\item \textbf{add\_product\_review.} 
Designed to simplify the process of adding a product review, the only parameter this function requires is the review text. We assume the review is posted on the website once this function is successfully invoked.
\item \textbf{respond.}
This function allows the agent to engage in dialogue with the user, enabling clarification or gathering of additional information.
\item \textbf{stop.}
The stop function signals the termination of the current task. 
When invoked, it indicates that the agent decided to end the task, and no further actions are required.
\end{itemize}

\subsection{Evaluation}\label{sec:evaluation}

To thoroughly evaluate the capabilities of Web agents, we established two distinct evaluation tracks: the single-turn track and the multi-turn track.

\vspace{3pt}
\noindent$\bullet$ \textbf{Single-turn track.}
In this track, the agent is given only one opportunity to execute the user's instruction.
The Web agent is expected to invoke the appropriate Web functions and deliver accurate results by configuring these functions with optimal parameters. Therefore, we define two metrics as follows:

\setlength{\leftmargini}{10pt} \begin{itemize}
\item \textbf{Function accuracy} (function acc): 
This metric assesses the agent's ability to select the correct function and provide parameters in the correct format. 
If the agent selects the appropriate tool for the task and the input parameters are correctly formatted, it receives a score of 1 for that task; otherwise, the score is 0.

\item \textbf{Result accuracy} (res acc):
For search and recommendation instructions, we leverage the rank \(r\) of the target product within the returned product list from the tool as the metric, formulated as: 
\begin{equation} \label{eqn1}
\text{Res Acc} = 
\begin{cases} 
1 - \frac{r - 1}{10}, & \text{if } r \leq 10, \\ 
0, & \text{if } r > 10.
\end{cases}
\quad \text{with } r \in \mathbb{N^+} 
\end{equation}
For review instructions, we assess the similarity between the agent's generated review and the user's actual review. 
We employ the sentence-transformer~\cite{reimers-2019-sentence-bert} model to compute the cosine similarity, yielding a res acc between 0 and 1.

\end{itemize}

\noindent$\bullet$ \textbf{Multi-turn track.}
\label{multiturn eval}
We believe it is crucial for Web agents to interact with users to receive feedback and continuously adjust their actions. 
Since using real humans for benchmark evaluation is impractical, we conduct \textbf{user simulators} based on LLMs to give real-time feedback.
Specifically, we provide the LLM with user profiles, target product information, or ground-truth reviews to facilitate high-quality interactions between user simulators and Web agents. Please refer to Figure~\ref{app: prompt user simulation} for the details of the user simulator prompt.

In addition to the two metrics used in the single-turn track, we introduce an additional evaluation metric: \textbf{average steps}. 
This metric measures efficiency by counting the total number of actions taken to complete the task, encouraging the agent to accomplish users' tasks with minimal attempts.
\section{Benchmark Analysis}\label{sec:benchmark_analysis}

\subsection{Statistic Analysis}

\begin{table}[]
\centering
\caption{Statistics of the PersonalWAB Benchmark.}
\vspace{-1em}
\begin{tabular}{cccc}
\hline
                              & items                   & Train        & Test       \\ \hline
\multirow{4}{*}{User}        & \# Users                     & 939          & 1,000      \\
                              & \# Avg. profile tokens  & \multicolumn{2}{c}{247}   \\
                              & \# Avg. behavior length & 32         & 38        \\
                              & \# Avg. behavior tokens & 7,597      &  9,270       \\ \hline 
\multirow{2}{*}{Instruction} & \#  Instructions                   & 6,896        & 2,174      \\
                              & \# Avg. tokens          & 46           & 45         \\ \hline
\multirow{2}{*}{Product}     & \#  Products                   & \multicolumn{2}{c}{8,236} \\
                              & \#  Avg. tokens         & \multicolumn{2}{c}{665}   \\ \hline
\end{tabular}
\vspace{-1em}
\label{tab: overall statistic}
\end{table}

We present the basic statistical information of our conducted PersonalWAB in Table~\ref{tab: overall statistic}. Since user profiles are generated in our benchmark, we analyze the diversity of all users and the consistency of each user's profile, to verify the reliability of PersonalWAB.

\begin{figure}[]
\setlength{\abovecaptionskip}{0.1cm}
\setlength{\belowcaptionskip}{0cm}
\centering
\includegraphics[width=1\linewidth]{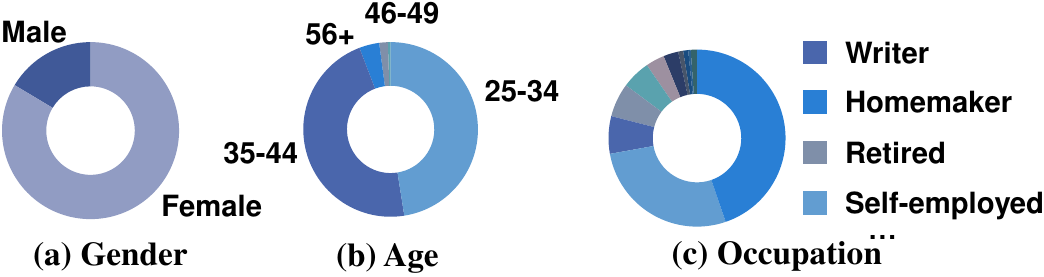}
\caption{Distribution of users by gender, age, and occupation.}
\vspace{-1em}
\label{fig: profile basic distri}
\end{figure}

\begin{figure}[]
\setlength{\abovecaptionskip}{0.1cm}
\setlength{\belowcaptionskip}{0cm}
\centering
\includegraphics[width=1\linewidth]{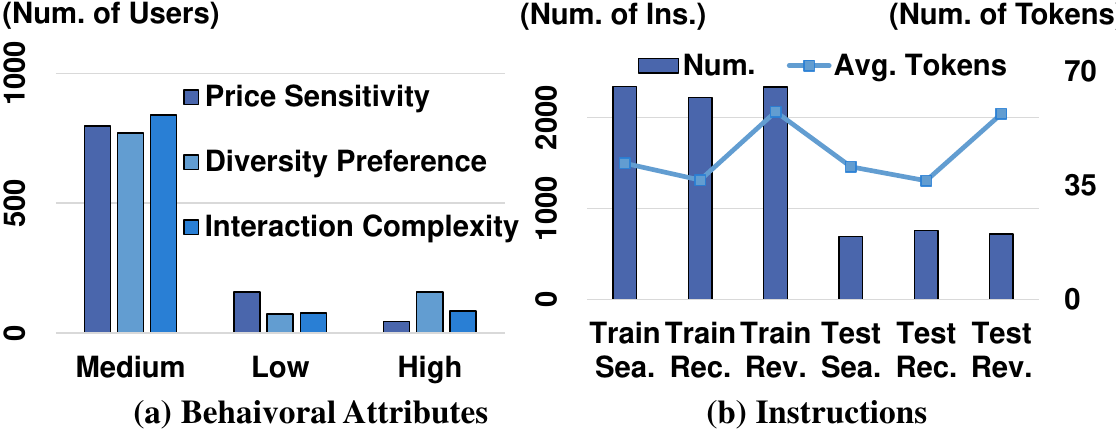}
\caption{ (a) Distribution of behaviors by Price Sensitivity, Diversity Preference, and Interaction Complexity; (b) Statistics of the instructions on different tasks.}
\vspace{-1em}
\label{fig: profile pref}
\end{figure}

\vspace{3pt}
\noindent$\bullet$ \textbf{User statistics.}
In Figure~\ref{fig: profile basic distri}, we present the basic attributes of user profiles to illustrate the distribution. The data shows a reasonable spread across gender and age groups, while occupation categories cover a wide range of fields, ensuring diverse professional backgrounds in the dataset.
The statistics in Figure~\ref{fig: profile pref} (a) highlight additional diversity in behavioral attributes such as \textit{Price Sensitivity}, \textit{Diversity Preference}, and \textit{Interaction Complexity}. 
Most users fall into the ``medium'' category across these behavioral aspects, and the ``high'' and ``low'' categories are less frequent, which allows for testing both typical and edge cases in personalized tasks. 

\vspace{3pt}
\noindent$\bullet$ \textbf{Instruction statistics.}
We examined the number and average token length of different instructions in Figure~\ref{fig: profile pref} (b). It is observed that the recommendation instructions have the smallest number of tokens because the recommendation is an exploratory task and doesn’t contain many user information requirements. The review instructions show slightly higher complexity than search and recommendation instructions, as they include many words for users to express their initial evaluations.

\subsection{Profile Consistency Evaluation}
\textbf{} 
\begin{figure}[]
\setlength{\abovecaptionskip}{0.1cm}
\setlength{\belowcaptionskip}{0cm}
\centering
\includegraphics[width=0.85\linewidth]{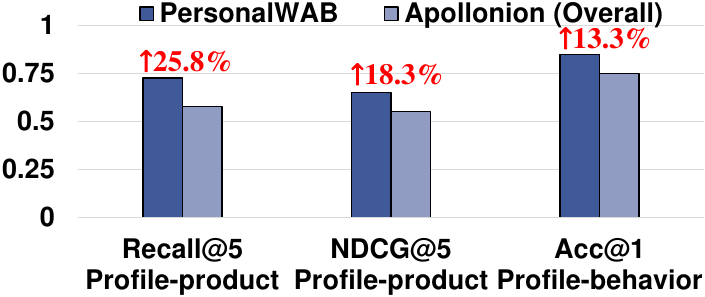}
\caption{Results of profile consistency evaluation experiments. Our generated profiles align better with users' actual Web behaviors and interested products than Apollonion~\cite{chen2024apollonion}.}
\vspace{-1em}
\label{fig: profile_eval}
\end{figure}
Following~\cite{chen2024apollonion},  we conducted experiments on profile-behavior consistency and profile-product consistency to verify how well the generated profiles align with users' actual past Web behaviors and potentially interesting products.
1) \textbf{Profile-behavior consistency evaluation.} 
The task was to match a user profile with the user's past Web behaviors among those of other users.
2) \textbf{Profile-product consistency evaluation.} 
The task involved ranking a mixture of previously interacted (positive) and randomly sampled (negative) items for a user, based on their profile.
The results in Figure~\ref{fig: profile_eval} show that our constructed profiles provide substantial improvements in both profile-product and profile-behavior consistency evaluations compared to Apollonion~\cite{chen2024apollonion}, showcasing the enhanced distinctiveness and alignment of the profiles with the actual user behaviors.
More details are in \S~\ref{app: profile_eval}.
\section{PUMA Framework}

\begin{figure}[t]
\setlength{\abovecaptionskip}{0.1cm}
\setlength{\belowcaptionskip}{0cm}
\centering
\includegraphics[width=1\linewidth]{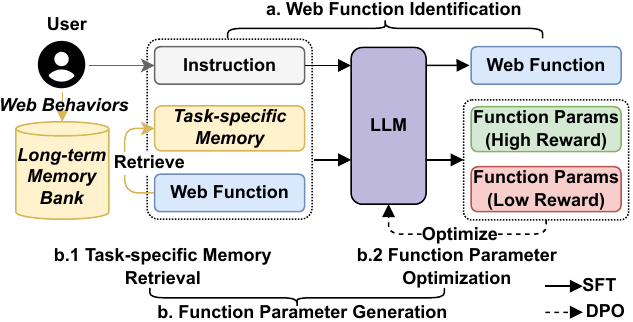}
\caption{Illustration of the PUMA framework, consisting of two main steps: Web Function Identification and Parameter Generation, which includes Task-specific Memory Retrieval and Function Parameter Optimization.}
\vspace{-1.5em}
\label{fig: method}
\end{figure}

To enable the Web agent to effectively complete tasks following user instructions, we propose a novel PUMA framework, which consists of two key steps: Web function identification and function parameter generation (see Figure~\ref{fig: method}).
First, we fine-tune an LLM (\eg LLaMa-2-7b~\cite{touvron2023llama}) with ``instruction-function'' pairs in the training set to identify the correct Web functions given a user instruction.
Then, we generate the appropriate parameters for the identified function.
To achieve this, PUMA first adopts a memory bank to store the users' long-term Web behaviors and utilizes a task-specific retrieval strategy to obtain the most relevant ones from the memory bank.
With the obtained user behaviors and features, we instruct the LLM to generate the corresponding parameters.
However, generating the appropriate parameters for the identified function poses a significant challenge due to the vast parameter space. 
To address this challenge, PUMA applies heuristic methods to construct pseudo-labels to further fine-tune the LLM and optimize parameter generation using DPO~\cite{2023dpo}, ensuring superior alignment with user preferences.

\subsection{Task-specific Memory Retrieval}
\noindent$\bullet$ \textbf{Long-term memory bank.}
The long-term memory bank is a storage system where we maintain a detailed record of each user's historical Web behaviors.
For a user $u$, we store detailed information about their purchased products $h_{purchase}$ and the associated reviews $h_{review}$, collectively denoted as $m$.
Specifically, the product details include attributes such as ``title'', ``price'', ``store'', and other relevant metadata, while the review details encompass the ``rating'', ``review title'', and ``comment'' provided by the user.
Formally, if the user $u$ has purchased $ n $ products, the long-term memory $ M $ is represented as: $M = \{ m_i \mid i = 1, 2, ..., n \}.$

\vspace{3pt}
\noindent$\bullet$ \textbf{Task-specific memory retrieval strategy.}
The task-specific memory retrieval strategy is designed to extract relevant information from the long-term memory bank based on the user's current instruction and the identified function.
When the user $u$ provides an instruction $i$ and the Web function $f$ is determined, we first retrieve the top $K$ memory entries by computing the cosine similarity between the instruction $i$ and each memory $m_j$ in the bank $M$. 
Then, based on the specific function $f$, we extract more targeted features from the retrieved memory.
1) If the Web function is related to search, we extract product details including the ``product title'', ``category'', ``price'', and ``store''. 
2) If the function pertains to the recommendation, we retain the product ``title'', ``category'', and ``parent ASIN'' (product ID). 
3) For review functions, only the user's past ratings and comments are kept.
This process can be formally defined as:
\begin{equation} \label{eqn2}
M_{i} =  \operatorname{Extract} \left( \operatorname{TopK} \left( { M }, \operatorname{sim}(i,  m_{j}), K \right), f \right).
\end{equation}
$ M_{i} $ represents the task-specific memory constructed for instruction $i$.
$ \operatorname{Extract}(\cdot, f)$ represents extracting targeted features based on the identified Web function $f$.
The similarity $\operatorname{sim}(i,  m_{j})$ is the cosine similarity between the instruction $i$ and memory entry $m_j$.

\subsection{Function Parameter Optimization}
Given the task-specific memory \( M_{i} \), the next step involves utilizing this memory to generate the Web function parameters.
We begin by using SFT to equip the model with a foundational ability to generate reasonable parameters.

\vspace{3pt}
\noindent$\bullet$ \textbf{Heuristic fine-tuning for parameter generation.}
The inputs for SFT are structured as a combination of the user instruction $i$, the task-specific memory $M_i$, and the identified Web function $f$. 
The labels are the Web function parameters, constructed using heuristic methods tailored to each Web function.
1) For the search function, we leverage ChatGPT~\cite{openai_gpt-4_2024} to generate textual queries based on the instruction, and memory.
2) For recommendations, the output consists of the most recent product ASINs from the same category found in the memory $M_i$.
3) For review functions, we use the actual review text provided by the dataset as the labels.
These heuristics help construct meaningful pseudo-labels for parameter generation, ensuring that the model learns to generate function parameters that are plausible and contextually appropriate for each Web function.

\vspace{3pt}
\noindent$\bullet$ \textbf{Diverse parameter sampling for pair-wise optimization.}
After SFT equips the model with fundamental ability, we further enhance the model's performance through DPO~\cite{2023dpo} over diverse parameter candidates.
We first generate a diverse set of function parameters with high-temperature sampling and beam search. 
These candidate parameters are then evaluated based on their result accuracy for instruction completion.
For instruction $i$, we collect best-performing (\(p_{i}^{\text{b}}\)) and worst-performing (\(p_{i}^{\text{w}}\)) parameter pairs to construct the pair-wise preference data, which is formally defined as \(\mathcal{D}_{\text{DPO}}\): 
\begin{equation} \label{eqn3}
\mathcal{D}_{\text{DPO}} = \left\{ \left( p_{i}^{\text{b}}, p_{i}^{\text{w}}, x_{i} \right) \right\},
\end{equation}
where \(x_{i}\) represents the input, which includes the user instruction $i$, task-specific memory $M_i$, and Web function $f$.

We then apply DPO to optimize the fine-tuned model \(\pi_{\text{ref}}\) by encouraging it to generate function parameters similar to \(p_{i}^{\text{b}}\) and discouraging it from generating function parameters similar to \(p_{i}^{\text{w}}\). 
The DPO loss is given by:
\begin{equation} \label{eqn4}
\mathcal{L}_{\text{DPO}} = -\mathbb{E} \left[ \log \sigma \left( \beta \log \frac{\pi_{\theta}(p^{\text{b}} \mid x)}{\pi_{\text{ref}}(p^{\text{b}} \mid x)} - \beta \log \frac{\pi_{\theta}(p^{\text{w}} \mid x)}{\pi_{\text{ref}}(p^{\text{w}} \mid x)} \right) \right],
\end{equation}
where \(\sigma(\cdot)\) is the sigmoid function, and \(\beta\) is a temperature-like parameter that controls the sensitivity of the model's preference to the log-ratio difference between the policy model \(\pi_{\theta}\) for optimization and reference model \(\pi_{\text{ref}}\) derived from the SFT stage.

\begin{table*}[]
\centering
\caption{Performance comparison between our proposed method, PUMA, and baselines in single-turn track. Bold numbers indicate the best performance in each column, while underlined numbers indicate the second-best performance.}
\vspace{-1em}
\setlength{\tabcolsep}{1.8mm}{
\resizebox{1\textwidth}{!}{
\begin{tabular}{cccccccccc}
\hline
\multicolumn{2}{c}{\multirow{2}{*}{Method (backbone)}} & \multicolumn{2}{c}{Search} & \multicolumn{2}{c}{Recommendation} & \multicolumn{2}{c}{Review} & \multicolumn{2}{c}{Overall} \\ \cline{3-10}
        && Function Acc.& Res Acc& Function Acc.& Res Acc& Function Acc.& Res Acc& Function Acc.& Res Acc\\ \hline
\multicolumn{2}{c}{No Memory (gpt-4o)} & \textbf{1.000}    & 0.647   & 0.092    & 0.000   & \textbf{1.000}    & 0.444   & 0.684    & 0.355   \\
\multicolumn{2}{c}{Random Memory   (gpt-4o)} & 0.974    & 0.640   & 0.296    & 0.018   & \underline{0.996}    & 0.442   & 0.745    & 0.357   \\
 \multicolumn{2}{c}{Last Memory  (gpt-4o)} & 0.937    & 0.626   & 0.432    & 0.028   & \textbf{1.000}    & 0.442   & 0.782    & 0.357   \\
   \multicolumn{2}{c}{Relevant Memory  (gpt-4o)}& 0.928    & 0.622   & 0.492    & 0.030   & \textbf{1.000}    & 0.443   & 0.800    & 0.356   \\
\multicolumn{2}{c}{ReAct~\cite{yao2023react}  (gpt-4o)} & 0.903    & 0.605   & 0.560    & 0.027   & \underline{0.996}    & 0.444   & 0.815    & 0.350   \\
  \multicolumn{2}{c}{RecMind~\cite{wang-etal-2024-recmind}  (gpt-4o)} & 0.981    & 0.645   & 0.226    & 0.017   & 0.990    & 0.442   & 0.721    & 0.359   \\ \hline
\multicolumn{2}{c}{PUMA(gpt-4o) }& \textbf{1.000}    & \underline{0.649}   & \underline{0.939}    & \underline{0.048}   & \textbf{1.000}    & \underline{0.449}   & \underline{0.979}    & \underline{0.373}   \\
\multicolumn{2}{c}{PUMA( LLaMA-7B  ) }   & \underline{0.996}    & \textbf{0.652}   & \textbf{0.987}    & \textbf{0.054}   & \textbf{1.000}    & \textbf{0.538}   & \textbf{0.994}    & \textbf{0.406}   \\ \hline
\end{tabular}}}
\vspace{-0.5em}
\label{tab: singleturn}
\end{table*}

\begin{table*}[]
\centering
\caption{Performance comparison between our proposed method, PUMA, and baselines in multi-turn track. F. Acc. represents function accuracy, R. Acc. stands for result accuracy, and Avg. Steps indicate the average number of steps taken by the agent to complete each instruction.}
\vspace{-1em}
\setlength{\tabcolsep}{1mm}{
\resizebox{1\textwidth}{!}{
\begin{tabular}{ccccccccccccc}
\hline
\multirow{2}{*}{Method (backbone)}    & \multicolumn{3}{c}{Search} & \multicolumn{3}{c}{Recommendation} & \multicolumn{3}{c}{Review} & \multicolumn{3}{c}{Overall} \\ \cline{2-13} 
                & F. Acc.& R. Acc.& Avg. Steps & F. Acc.& R. Acc.& Avg. Steps & F. Acc.& R. Acc.& Avg. Steps &F. Acc.& R. Acc.& Avg. Steps \\ \hline
No Memory (gpt-4o)       & 0.996  & 0.656  & 2.398 & 0.096  & 0.000  & 2.420 & \textbf{1.000}  & 0.446  & 2.019 & 0.685  & 0.358  & 2.280 \\
Random Memory (gpt-4o)   & \underline{0.999}  & 0.680  & 4.193 & 0.703  & 0.042  & 4.474 & \textbf{1.000}  & 0.448  & 2.007 & 0.896  & 0.380  & 3.564 \\
Last Memory (gpt-4o)    & 0.996  & 0.676  & 4.229 & 0.708  & \underline{0.045}  & 4.252 & \textbf{1.000}  & 0.449  & 2.007 & 0.897  & 0.381  & 3.498 \\
Relevant Memory (gpt-4o) & 0.996  & \underline{0.686}  & 4.233 & \underline{0.715}  & 0.042  & 4.564 & \underline{0.999}  & 0.448  & 2.008 & \underline{0.899}  & \underline{0.383}  & 3.609 \\
ReAct~\cite{yao2023react} (gpt-4o)          & 0.996  & 0.674  & 4.657 & 0.218  & 0.013  & 5.468 & 0.974  & 0.448  & 2.129 & 0.718  & 0.369  & 4.098 \\
Reflexion~\cite{2024reflexion} (gpt-4o)     & \textbf{1.000}  & \underline{0.686}  & 5.406 & 0.281  & 0.014  & 6.145 & 0.976  & 0.449  & 2.145 & 0.741  & 0.373  & 4.579 \\
RecMind~\cite{wang-etal-2024-recmind} (gpt-4o)        & 0.997  & 0.642  & 6.728 & 0.347  & 0.026  & 6.003 & 0.997  & \underline{0.451}  & 2.107 & 0.771  & 0.364  & 4.938 \\
InteRecAgent~\cite{huang2023interectagent} (gpt-4o)   & \underline{0.999}  & 0.642  & 3.110 & 0.618  & 0.022  & 3.008 & \textbf{1.000}  & 0.447  & 2.001 & 0.867  & 0.362  & 2.706 \\ \hline
PUMA (gpt-4o) & \underline{0.999}   & \textbf{0.720}   & 5.082  & \textbf{0.984}      & \textbf{0.052}     & 3.791     & \textbf{1.000}   & \textbf{0.453}   & 2.002  & \textbf{0.994}   & \textbf{0.399}   & 3.608   \\ \hline
\end{tabular}}}
\label{tab: multiturn}
\vspace{-0.5em}
\end{table*}

\section{Experiments}

We evaluate a range of baselines that employ different strategies for selecting and utilizing user history. 
These baselines are categorized into three groups: Memory Retrieval based Methods (No Memory, Random Memory, Last Memory, Relevant Memory), Enhanced Reasoning Methods (ReAct~\cite{yao2023react}, Reflexion~\cite{2024reflexion}), and Recommendation-Specific Memory Frameworks (Recmind~\cite{wang-etal-2024-recmind}, InteRecAgent~\cite{huang2023interectagent}).
The implementation details of baselines and our method are illustrated in \S~\ref{app: baseline} and \S~\ref{app: implementation}. All methods use the same prompt template, with differences only in the memory component. The detailed prompt design is provided in Figure~\ref{app: single prompt} and Figure~\ref{app: multi prompt}.

\subsection{Main Results}
We evaluated baselines and our framework on both single-turn and multi-turn evaluation tracks. 
\subsubsection{Single-turn Track}
The results on the single-turn track are shown in Table~\ref{tab: singleturn}, highlighting several key insights: 1) It was observed that while search instructions had high function accuracy, the performance for recommendation instructions was poor. Further analysis revealed that many recommendation instructions were incorrectly assigned to the search function, as visualized in Figure~\ref{fig: multi_flow} (b), indicating the great difficulty in function selection. 
2) Methods incorporating relevant memory and ReAct show improved function accuracy, suggesting that retrieving relevant information and incorporating reasoning improve function selection.
3) The result accuracy for all baselines remains similar to the naive ``No Memory'' baseline, implying these methods fail to significantly enhance personalized task execution.
4) In contrast, PUMA achieves the highest function accuracy across tasks, with task-specific memory enabling the agent to focus on relevant behaviors and features, leading to higher result accuracy. 
Additionally, PUMA delivers the best overall performance while using shorter memory and a smaller LLM, highlighting the efficiency and effectiveness of our approach.

\subsubsection{Multi-turn Track}
The multi-turn track results (Table~\ref{tab: multiturn}) offer valuable insights into how different methods handle complex interactions.
1) First, baselines perform better in search and recommendation tasks compared to the single-turn track, benefiting from multiple attempts and user feedback, while review tasks show minimal improvement, as the agents typically follow a straightforward flow with limited feedback opportunities.
2) The memory retrieval baselines follow similar trends to the single-turn track, with relevant memory improving function accuracy and result accuracy, but at the cost of additional steps. 
3) ReAct and Reflexion perform worse than memory retrieval methods, requiring more steps and yielding lower accuracy. 
The complexity of these methods, which include reasoning and self-reflexion, seems to hinder task efficiency and accuracy with extra input token length.
4) RecMind also requires a higher number of steps, as it performs additional function calls, but struggles with instruction identification. 
InteRecAgent uses fewer steps due to its streamlined memory, but this simplification results in lower result accuracy. 5) Our Task-specific Memory method performs strongly, particularly in search and recommendation tasks. 
By extracting relevant information and filtering out redundant data, it enables more informed decisions with fewer steps. 
Although we did not evaluate the full PUMA approach due to model limitations in multi-turn settings, the results highlight the importance of task-specific memory in enhancing both efficiency and accuracy.

\subsection{In-depth Analysis}
We performed a comprehensive analysis of PUMA, including experiments on ablation study, memory length, efficiency, action transitions, multi-turn performance variation, function usage and outcome accuracy, search function implementation, and zero-shot and few-shot performance. Due to space limitations, we present the results for efficiency in this section, while the remaining analyses are provided in \S~\ref{app:more analysis}.

\subsubsection{Analysis on efficiency.}
\begin{figure}[t]
\setlength{\abovecaptionskip}{0.1cm}
\setlength{\belowcaptionskip}{0cm}
\centering
\includegraphics[width=1\linewidth]{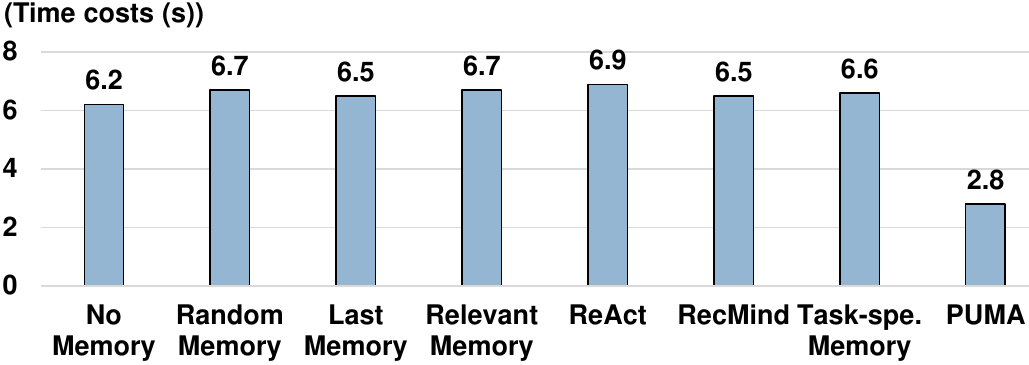}
\caption{Comparison between the average task completion time (in seconds) for different methods.}
\vspace{-2em}
\label{fig: effi}
\end{figure}

In real-world applications, task completion time is crucial for delivering a smooth user experience. 
To evaluate it, we measured the time taken to complete the user instruction in the single-turn track. 
Each method was tested on 100 randomly selected tasks, and the average completion time was calculated. The results in Figure~\ref{fig: effi} show that GPT-based methods have similar completion times, ranging from 6.5 to 6.9 seconds, due to memory processing overhead.
In contrast, our PUMA framework significantly outperforms all baselines, with an average time of 2.8 seconds. 
This efficiency gain stems from PUMA’s smaller model and compact memory structure, minimizing inference time, making it highly effective for real-world Web applications where quick response times are essential.

\section{Conclusion and Future Work}
In this paper, we advanced general LLM-based Web agents into the era of personalized Web agents, aiming to offer users tailored and customized services. 
We formulated the task of LLM-empowered personalized Web agents and identified the goal of leveraging personalized user data to achieve personalized instruction understanding and action execution (Web function call).  
For training and evaluation, we constructed the first PersonalWAB benchmark on three personalized Web tasks. 
We proposed PUMA, a novel personalized alignment framework with task-specific memory and function parameter optimization strategies, to adapt LLMs to personalized Web agents. 
Extensive experiments on PersonalWAB demonstrate that PUMA consistently surpasses existing Web agents, aligning better with personalized user instructions and preferences. We believe that the task, benchmark, and framework for LLM-empowered personalized Web agents will broaden the research scope, introduce new challenges, and inspire novel methods in Web agent scenarios.

While our research lays the groundwork for personalized Web agents, several avenues for future exploration remain. 
First, we plan to extend PersonalWAB by incorporating more diverse task scenarios to further challenge and evaluate Web agents' personalization capabilities. 
Second, integrating more sophisticated user modeling techniques, such as dynamic preference learning, could enhance agents' adaptability to evolving user needs. 
Third, exploring user-in-the-loop settings presents an exciting opportunity to improve task execution by actively involving users in the process. 
This includes developing agents that can better integrate user feedback, proactively identify missing information, and engage with users to request necessary details, thereby enhancing the overall effectiveness and efficiency of task completion.

Although this work makes significant strides in personalized Web agents, it is important to note some limitations. Ethical and privacy considerations, along with the scope of our study, warrant further discussion. These aspects are elaborated in \S~\ref{discussion}, where we provide a detailed exploration of challenges associated with user data usage, potential biases in personalization, and the current focus on shopping scenarios. While our approach demonstrates strong adaptability, extending it to broader Web environments presents additional complexities that require further investigation.


\begin{acks}
The work described in this paper was supported by the Research Grants Council of Hong Kong (PolyU/15209724, PolyU/15207821, PolyU/15207122, PolyU/15213323) and PolyU internal grants (BDWP).
\end{acks}

\bibliographystyle{ACM-Reference-Format}
\balance
\bibliography{sample-base}

\appendix
\section{Details}

\subsection{Details of Profile Evaluation Experiments}
\label{app: profile_eval}
\textbf{Profile-behavior consistency evaluation.}
Given a specific user profile, the task is to identify the correct user from a group of candidate users, consisting of the true user and several negative users.
Each candidate user is represented by their behavior sequence, and the objective is to determine which candidate aligns best with the given profile. 
The evaluation metric used is top-1 accuracy, indicating the ability of the profile to distinctly and accurately match the correct user based on their behaviors.

\begin{table*}[t]
\centering
\caption{Alation study on key components of PUMA in single-turn track.}
\vspace{-1em}
\setlength{\tabcolsep}{1mm}{
\resizebox{1\textwidth}{!}{
\begin{tabular}{ccccccccc}
\hline
\multirow{2}{*}{Method} & \multicolumn{2}{c}{Search} & \multicolumn{2}{c}{Recommendation} & \multicolumn{2}{c}{Review} & \multicolumn{2}{c}{Overall} \\ \cline{2-9} 
                          & Function Acc & Result Acc & Function Acc & Result Acc & Function Acc & Result Acc & Function Acc & Result Acc \\ \hline
PUMA             & 0.996    & 0.652      & 0.987    & 0.054      & 1.000    & 0.538      & 0.994    & 0.406      \\ \hline
w/o Task-specific Memory                & 0.990    & 0.643      & 0.992    & 0.008      & 1.000    & 0.496      & 0.994    & 0.373      \\
w/o SFT & 1.000    & 0.000      & 0.983    & 0.000      & 1.000    & 0.160      & 0.994    & 0.054      \\
w/o DPO                   & 0.996    & 0.648      & 0.987    & 0.047      & 1.000    & 0.529      & 0.994    & 0.399      \\ \hline
\end{tabular}}}
\vspace{-1em}
\label{tab: ablation}
\end{table*}

\begin{table*}[t]
\centering
\caption{Performance comparison of different memory token lengths in PUMA.}
\vspace{-1em}
\setlength{\tabcolsep}{1mm}{
\resizebox{1\textwidth}{!}{
\begin{tabular}{ccccccccc}
\hline
\multirow{2}{*}{Memory Length} & \multicolumn{2}{c}{Search} & \multicolumn{2}{c}{Recommendation} & \multicolumn{2}{c}{Review} & \multicolumn{2}{c}{Overall} \\ \cline{2-9} 
                               & Function Acc   & Result   Acc  & Function   Acc      & Result   Acc     & Function   Acc  & Result   Acc & Function   Acc  & Result   Acc  \\ \hline
256 & 0.997 & 0.651 & 0.985 & 0.019 & 1.000 & 0.530 & 0.994 & 0.391 \\
512 & 0.991 & 0.648 & 0.988 & 0.032 & 1.000 & 0.531 & 0.993 & 0.395 \\
768 & 0.996 & \textbf{0.652} & 0.987 & \textbf{0.054} & 1.000 & \textbf{0.538} & 0.994 & \textbf{0.406} \\ \hline
\end{tabular}}}
\vspace{-1em}
\label{tab: mem length}
\end{table*}

\textbf{Profile-product consistency evaluation.}
In this task, a given user profile is used to rank a set of candidate items, which include a mixture of positive items (previously interacted with by the user) and negative items (randomly sampled from an item pool). 
The objective is to prioritize positive items over negative items, leveraging the user profile for accurate ranking. 
The task is evaluated with NDCG@5 and Recall@5, which measure the profile's ability to reflect the user's preferences.

We adopt the same setting with~\cite{chen2024apollonion}, set the number of positive samples to 1 and 3, and negative samples to 4 and 7 in the user prediction and recommendation tasks, respectively. To ensure a fair evaluation, both experiments are conducted in separate, isolated sessions. 
Experiments are conducted with \texttt{gpt-4o-mini-2024-07-18}, and results in Figure~\ref{fig: profile_eval} show our profile exhibits significant improvements across both tasks, indicating that our profiles are more distinct and better aligned with user behaviors.

\subsection{Details of Baselines}
\label{app: baseline}
We evaluate a range of baselines that employ different strategies for selecting and utilizing user Web behavioral history. 
These baselines are categorized into three groups: Memory Retrieval Methods, Enhanced Reasoning Methods, and Recommendation-Specific Memory Frameworks.

\textbf{Memory Retrieval Methods.}
We include various simple memory mechanisms as baselines, aiming to explore different strategies for selecting and utilizing user history. This helps us understand how each memory selection technique impacts task performance.
1) No Memory: The agent performs tasks without accessing any user history, relying solely on the current instruction. 
2) Random Memory: This approach randomly selects behaviors from the user's history. 
3) Last Memory: Uses only the most recent behaviors from the user's history, focusing on the assumption that recent context is more relevant for current instruction. 
4) Relevant Memory: Selects past behaviors based on cosine similarity with the current instruction, aiming to filter out the most contextually relevant details for the task.

\textbf{Enhanced Reasoning Methods.}
We also tested frameworks designed to enhance the agent’s reasoning and decision-making capabilities.
1) ReAct: Proposed by~\cite{yao2023react}, this framework instructs the language model to think before taking an action and generate ``Thought: {some reasoning} Action: {some JSON format action argument}'' to interact, enabling the model to deliberate over the information available and decide on the most appropriate action.
2) Reflexion: Reflexion~\cite{2024reflexion} builds on frameworks of ReAct by adding a self-evaluation phase, where the agent reviews and analyzes its previous actions and outcomes before proceeding. 
This process allows the agent to recognize mistakes, reassess decisions, and refine its strategy in subsequent interactions.
We tested this baseline only in the multi-turn track, where the agent treats each user message as feedback for Reflexion and adjustment.

\textbf{Recommendation-Specific Memory Frameworks.}
Recommendation tasks are inherently personalized, as they rely on a deep understanding of user preferences and behaviors~\cite{2018sasrec,huang2023interectagent,wang-etal-2024-recmind,gao2023cirs,ADDRL2024MM}. 
Given this, we include baselines that leverage memory mechanisms developed for recommendation agents, assessing their ability to enhance personalization in our context.
1) RecMind: An LLM-powered agent designed for general recommendation purposes~\cite{wang-etal-2024-recmind}, consists of two parts of memory, personalized memory, and world knowledge. 
Personalized Memory includes individualized user information, such as their reviews or ratings for a particular item.
World Knowledge consists of item metadata information and real-time information accessed with a Web search function.
In our setup, we retain the personalized memory containing user reviews and ratings, and we incorporate an additional function to enable RecMind to access detailed product information.
2) InteRecAgent: Proposed by~\cite{huang2023interectagent}, this framework uses LLMs as the core reasoning engine while utilizing recommender models as functions for interactive recommendations.
Its memory structure includes a candidate bus (which stores current item candidates) and a user profile that captures three facets of user preferences: ``like'', ``dislike'', and ``expect''.
We adopt the user profile memory in our experiments and allow the agent to update this profile at the end of each task.
As the user profile is synthesized by LLMs based on conversation history with the user, we evaluate this method only in the multi-turn setting, where ongoing dialogue allows for continuous adaptation of the user profile.

\subsection{Implementation Details}
\label{app: implementation}
\textbf{Benchmark.}
We utilize \texttt{gpt-4o-mini-2024-07-18} for generating user profiles, as it excels at extracting detailed user preferences, particularly in capturing brand preferences.
For user instruction creation, we employ \texttt{claude-3-5-sonnet@20240620}, selected for its ability to produce instructions in a natural and human-like tone.
In multi-turn track, \texttt{gpt-4o-mini-2024-07-18} is also used to simulate user messages, as it follows the instructions better to give user messages.

\textbf{Baselines.}
We use \texttt{gpt-4o-mini-2024-07-18} as the base language model across all baseline methods. 
For memory retrieval baselines, we set the memory length to 50 behaviors for the single-turn track and 20 behaviors for the multi-turn track, allowing additional input length for user messages and function results. 
For the Relevant Memory method, we calculate cosine similarity using the sentence-transformer~\cite{reimers-2019-sentence-bert} to identify relevant behaviors.
The ReAct baseline is combined with the Last Memory approach to ensure that reasoning processes have recent context, and we further extend this with a Reflexion mechanism for multi-turn scenarios. 
For RecMind, the memory length is set to 400 behaviors, as it only contains user reviews and ratings, and we added an extra ``get\_product\_details\_by\_asin'' function for the agent to retrieve detailed product information.
In the InteRecAgent setup, we first construct the memory using historical behaviors and the training dataset before evaluating performance on the test set.

\textbf{PUMA.}
We generate the function parameters for search using \texttt{gpt-4o-mini-2024-07-18} as the initial SFT labels. 
In the function parameter optimization phase, we fine-tune the LLaMA2-7B~\cite{touvron2023llama} model with LoRA~\cite{hu2022lora} using 4 × 24GB NVIDIA A5000 GPUs. 
The learning rate is set to 4e-3 for the SFT and 5e-5 during the DPO stage, with a batch size of 1 per GPU. 
Due to the maximum sequence length we can afford during training is limited, we constrain the memory token length to 256, 512, and 768 tokens.
To generate diverse function parameters, we set a temperature of 1.5 to increase output variability and use a beam search with a beam size of 10.

\subsection{More Analysis}
\label{app:more analysis}
\subsubsection{Ablation study.}
We conducted an ablation study (see Table~\ref{tab: ablation}) to assess the impact of PUMA’s key components. 
First, removing the memory leads to a significant drop in result accuracy across all tasks, highlighting the importance of memory in retaining relevant information for function parameter generation.
Second, when memory is retained but the SFT phase is removed, result accuracy dramatically declines. 
This indicates that without fine-tuning, the model struggles to generate function parameters that align with user needs.
Finally, removing the DPO phase results in a slight performance decrease, suggesting that DPO plays a crucial role in aligning the model with user preferences, and improving the quality of function parameters.

\subsubsection{Analysis on memory length.}
We evaluated the impact of different memory token lengths on our framework's performance across tasks. 
The experiment measured both function accuracy and result accuracy with varying memory sizes.
The results in Table~\ref{tab: mem length} indicate that increasing memory length has minimal impact on function accuracy, with the model maintaining similar performance regardless of memory size. 
However, memory length has a significant effect on result accuracy, particularly in recommendation tasks. 
Shorter memory lengths reduce the number of stored products, limiting the model’s ability to select appropriate product IDs, which leads to a noticeable drop in recommendation accuracy.
In contrast, search and review tasks are less sensitive to memory length changes, as the agent relies more on information from the instruction rather than the memory. 
This reduced dependence on memory in these tasks may also limit the model’s potential to improve performance further.

\subsubsection{Analysis on action transitions.} 
\begin{figure}[t]
\setlength{\abovecaptionskip}{0.1cm}
\setlength{\belowcaptionskip}{0cm}
\centering
\includegraphics[width=1\linewidth]{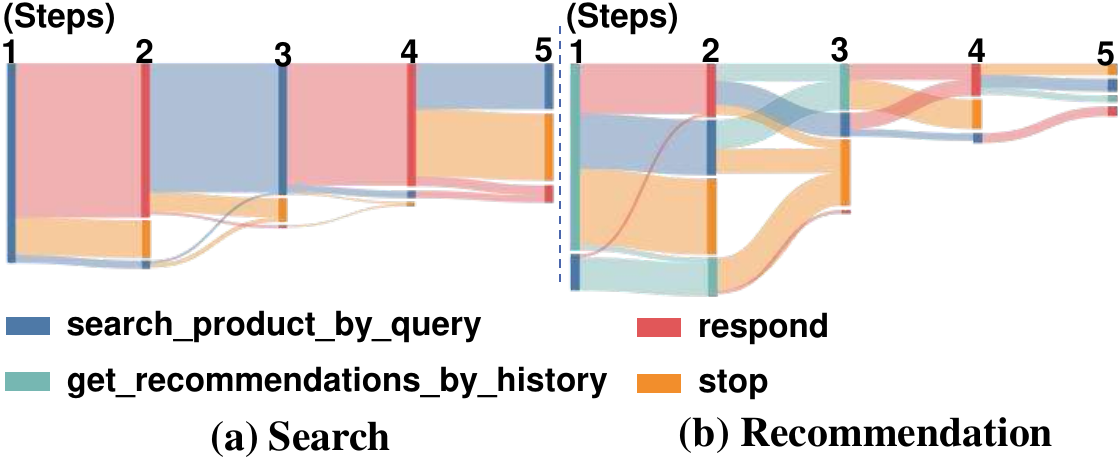}
\caption{Transitions of the agent’s actions in multi-turn search and recommendation tasks. Each color represents a specific function. The horizontal axis shows interaction steps, while the width of each color band indicates the proportion of the agent’s focus on that action. The flow between steps illustrates how the agent adapts its strategy over steps.}
\label{fig: multi_flow}
\end{figure}

We collected PUMA's actions in each interaction turn within the multi-turn track. Review instructions were removed, as the agent typically completes them in just two steps. The results, visualized in Figure~\ref{fig: multi_flow}, provided the following insights.
1) For search instructions, the agent tends to alternately call ``search'' and ``respond'' functions. It is reasonable as the agent could receive user feedback via the ``respond'' function and thus adjust its search action. 2) The bond for the recommendation instructions is more entangled in Figure~\ref{fig: multi_flow} (b), indicating a more complex action transition. This underlines the challenge in multi-turn recommendation tasks, where correctly identifying user intent and dynamically adjusting actions are more difficult than in straightforward tasks like search.

\begin{figure}[t]
\setlength{\abovecaptionskip}{0.1cm}
\setlength{\belowcaptionskip}{0cm}
\centering
\includegraphics[width=1\linewidth]{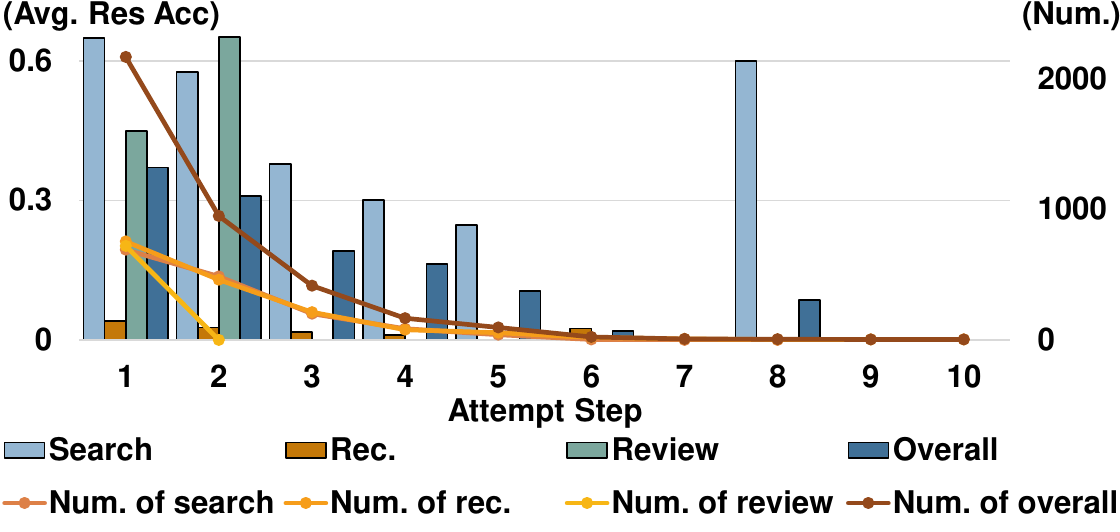}
\caption{Analysis of the agent's performance across multiple attempts in multi-turn track.}
\vspace{-0.5em}
\label{fig: multi_perfor}
\end{figure}

\subsubsection{Analysis of multi-turn performance variation.} We evaluated the agent’s performance over multiple attempts in the multi-turn track. For each instruction, we measured the ``Res Acc,'' and the number of solved tasks as the number of attempt steps increased. The results are shown in Figure~\ref{fig: multi_perfor}, and we had the following observations.
1)  The high task count within the first five attempts indicates that most tasks are completed early on. The ``review'' task is typically finished within the first two attempts, as there's little need for the agent to interact with users about the review requirements.
2) ``Res Acc'' is high during the initial attempts and declines with each subsequent attempt. This is because the easier tasks can be solved in a few attempts, leaving the more difficult tasks for later.
3) There are a few outliers where tasks achieve higher ``Res Acc'' in later steps. However, these cases are rare, involving only one or two tasks that result in the outlier.
4) The declined ``Res Acc'' also suggested that the agent struggles to leverage user feedback in later attempts effectively. This may be due to the lack of multi-turn training data, preventing us from tuning the agent accordingly.

\begin{table}[h]
\centering
\caption{Single-turn performance comparison of different methods in terms of function accuracy (F. Acc.), result accuracy (R. Acc.), and outcome accuracy (O. Acc.) in search and recommendation.}
\vspace{-1em}
\setlength{\tabcolsep}{1mm}{
\resizebox{0.48\textwidth}{!}{
\begin{tabular}{ccccccc}
\hline
\multirow{2}{*}{Method} & \multicolumn{3}{c}{Search} & \multicolumn{3}{c}{Recommendation} \\ \cline{2-7} 
                        & F. Acc. & R. Acc. & O. Acc. & F. Acc. & R. Acc. & O. Acc. \\ \hline
No Memory       & 1.000  & 0.647  & 0.647  & 0.092  & 0.000  & 0.155  \\
Random Memory   & 0.974  & 0.640  & 0.642  & 0.296  & 0.018  & 0.159  \\
Last Memory     & 0.937  & 0.626  & 0.632  & 0.432  & 0.028  & 0.161  \\
Relevant Memory & 0.928  & 0.622  & 0.631  & 0.492  & 0.030  & 0.159  \\
ReAct~\cite{yao2023react}          & 0.903  & 0.605  & 0.628  & 0.560  & 0.027  & 0.160  \\
RecMind~\cite{wang-etal-2024-recmind}        & 0.981  & 0.645  & 0.647  & 0.226  & 0.017  & 0.152  \\ \hline
PUMA  & \textbf{1.000}  & \textbf{0.649}  & \textbf{0.649}  & \textbf{0.939}  & \textbf{0.048}  & \textbf{0.164}  \\ \hline
\end{tabular}}}
\vspace{-2em}
\label{tab:occ}
\end{table}

\begin{table}[h]
\centering
\caption{Comparison of search result accuracy using BM25 and Dense retrieval methods in single-turn track.}
\vspace{-1em}
\begin{tabular}{ccc}
\hline
\multirow{2}{*}{Method} & \multicolumn{2}{c}{Search (Result Accuracy)} \\ \cline{2-3}
       & BM25  & Dense Retrieval\\ \hline
No Memory       & 0.647  & 0.502  \\
Random  Memory        & 0.640  & 0.504  \\
Last  Memory         & 0.626  & 0.498  \\
Relevant Memory       & 0.622  & 0.499  \\
ReAct~\cite{yao2023react}           & 0.605  & 0.496  \\
RecMind~\cite{wang-etal-2024-recmind}         & 0.645  & 0.498  \\ \hline
PUMA  & \textbf{0.649}  & \textbf{0.506}  \\ \hline
\end{tabular}
\vspace{-1em}
\label{tab:search_comparison}
\end{table}

\begin{table*}[]
\centering
\caption{Performance comparison in zero-shot and few-shot scenarios in single-turn track.}
\vspace{-1em}
\setlength{\tabcolsep}{1mm}{
\resizebox{1\textwidth}{!}{
\begin{tabular}{ccccccccc}
\hline
\multirow{2}{*}{Method} & \multicolumn{2}{c}{Search} & \multicolumn{2}{c}{Recommendation} & \multicolumn{2}{c}{Review} & \multicolumn{2}{c}{Overall} \\ \cline{2-9} 
                        & Function Acc.  & Result   Acc. & Function   Acc.     & Result   Acc.    & Function   Acc. & Result   Acc.& Function   Acc. & Result   Acc. \\ \hline
No Memory       & \textbf{1.000}  & 0.684  & 0.050  & 0.000  & \textbf{1.000}  & 0.388  & 0.625  & 0.328  \\
Random Memory   & 0.974  & 0.684  & 0.301  & 0.060  & 0.996  & 0.391  & 0.715  & 0.352  \\
Last Memory     & \textbf{1.000}  & 0.683  & 0.314  & 0.058  & \textbf{1.000}  & 0.396  & 0.730  & 0.353  \\
Relevant Memory & 0.928  & 0.675  & 0.405  & 0.078  & \textbf{1.000}  & \textbf{0.397}  & 0.743  & 0.358  \\
ReAct~\cite{yao2023react}           & 0.945  & 0.675  & 0.475  & 0.080  & 0.996  & 0.393  & 0.774  & 0.358  \\
RecMind~\cite{wang-etal-2024-recmind}         & 0.973  & 0.680  & 0.320  & 0.063  & 0.996  & 0.394  & 0.722  & 0.354  \\ \hline
PUMA  & \textbf{1.000}  & \textbf{0.686}  & \textbf{0.892}  & \textbf{0.090}  & \textbf{1.000}  & 0.396  & \textbf{0.958}  & \textbf{0.366}  \\ \hline
\end{tabular}}}
\vspace{-1em}
\label{tab:zero_few_shot_performance}
\end{table*}

\subsubsection{Analysis on function usage and outcome accuracy.}
In real-world applications, users care about whether the system retrieves relevant results rather than the specific function it employs. While search and recommendation serve different roles, their distinction may not always be meaningful from a user's perspective. For real users, the ultimate priority lies in the relevance of the returned items rather than the specific function used by the agent. 

To better align with this user-centric goal, we introduce Outcome Accuracy (O. Acc.), a metric that evaluates the correctness of the returned results independent of the function used for search and recommendation tasks. As shown in Table~\ref{tab:occ}, function accuracy (F. Acc.) and result accuracy (R. Acc.) vary significantly between search and recommendation tasks. However, Outcome Accuracy provides a more balanced perspective on the effectiveness of different methods. Our results show that PUMA achieves the highest Outcome Accuracy across both search and recommendation tasks, demonstrating its ability to deliver relevant results without being constrained by function selection. By incorporating this metric, we offer a more comprehensive evaluation that prioritizes relevance over strict function adherence, reflecting real-world user needs.

\subsubsection{Analysis on search function implementation.}

In our benchmark, the implementations of both the search and recommendation functions are flexible and not restricted to a specific method. Our focus is not on optimizing these functions but on evaluating the agent’s ability to effectively utilize the provided function in a personalized setting.  

We conducted an alternative retrieval experiment in the single-turn track, replacing BM25 with a dense retrieval model based on Sentence-BERT~\cite{reimers2019sentencebert}. The results in Table~\ref{tab:search_comparison} indicated that while dense retrieval captures richer semantic representations, it also introduces noise by embedding extensive product details, leading to a slight degradation in result accuracy across all baselines. And different retrieval methods may impact performance, but the overall trends remain consistent. Despite these variations, our PUMA framework consistently outperformed all baselines, regardless of the retrieval model used. Given the modular nature of our benchmark, future work could explore alternative retrieval models and different recommendation strategies, further analyzing how function-level improvements lead to final performance.

\subsubsection{Analysis on zero-shot and few-shot performance.}
Since our user sampling is fully random, some users naturally have limited historical data, leading to zero-shot and few-shot scenarios. To evaluate performance in such cases, we analyzed 139 users with fewer than 10 historical records (16.2\% of the test set) in the single-turn track. Results in Table~\ref{tab:zero_few_shot_performance} show task-dependent effects: Search performance remains stable or improves slightly due to reduced irrelevant information, while Recommendation performance also improves, as limited memory simplifies retrieval. However, Review tasks decline, as the lack of past reviews hinders personalized responses. Despite these variations, PUMA consistently outperforms all baselines, achieving the highest accuracy across tasks, particularly excelling in Recommendation scenarios. These results demonstrate that PUMA remains effective even in zero-shot and few-shot conditions, highlighting its adaptability when user history is sparse.

\section{Discussion}
\label{discussion}
\subsection{Ethical and Privacy Considerations}
The integration of personalized data in Web agents plays a crucial role in improving task efficiency and relevance. By leveraging user history, preferences, and past interactions, personalized Web agents can better align with individual needs and enhance task completion. However, the use of such data raises important ethical and privacy concerns that must be carefully considered.

One primary ethical concern is fairness in personalization. If personalization is not properly managed, it may introduce or amplify biases in decision-making processes. For example, recommendation algorithms might reinforce popularity bias~\cite{abdollahpouri2019unfairnesspopularitybiasrecommendation}, leading to a concentration of exposure on frequently suggested items while limiting diversity. Additionally, personalization can inadvertently discriminate against certain user groups if historical biases exist in the training data~\cite{2018equityof}. To mitigate these risks, fairness-aware personalization techniques and diversity-promoting strategies should be considered when designing personalized Web agents.

From a privacy perspective, handling user history introduces risks related to data security. Users may inadvertently disclose sensitive information, such as browsing behaviors, or purchase histories, which could be exploited if not properly secured. To ensure privacy protection, future work about personalized Web agents should incorporate privacy-preserving techniques, reducing the risk of data leakage while maintaining personalization benefits.

\subsection{Scope of This Work}
We chose the shopping domain as a representative and practically significant application to demonstrate the capabilities of personalized Web agents. Shopping platforms provide structured user behavior data and diverse interactions, such as search, recommendation, and product reviews, making them ideal testbeds for evaluating personalization strategies. Additionally, personalization in shopping directly impacts user experience and decision-making, highlighting the effectiveness of adaptive Web agents.

While our benchmark is designed on shopping-related Web functions, the framework introduced in this work is generalizable and can be extended to a broader range of Web applications, such as news recommendation, and social media content curation. These domains, like shopping, rely on user behavior modeling to enhance relevance and improve interaction quality.
\section{Prompting Details}
The prompt template for profile generation is shown in Figure~\ref{app: prompt profile generation} and Figure~\ref{app: prompt profile generation cho}.
The prompt template for instruction generation is shown in Figure~\ref{app: prompt search generation}, Figure~\ref{app: prompt rec generation}, and Figure~\ref{app: prompt review generation}.
The prompt template for the user simulator is shown in Figure~\ref{app: prompt user simulation}. Last, the prompt templates in task executing are shown in Figure~\ref{app: single prompt} and Figure~\ref{app: multi prompt}.
\begin{figure*}

\FancyBox{User Profile Generation}{
You will act as an online shopper.

Given your time-series historical purchase information and corresponding reviews, 
you need to summarize and choose the most accurate and relevant option best describing you.

During summarizing you should obey the following procedures:

First, summarize your basic information, and choose or fill the most accurate and relevant option for each category.

The categories and options are as follows:
\setlength{\leftmargini}{10pt} \begin{itemize}

\item\textbf{Gender}: <GENDER>.

\item\textbf{Age}: <AGE>.

\item\textbf{Occupation}:<OCCUPATION>.

\item\textbf{Price Sensitivity}: <PRICE SENSITIVITY>.

\item\textbf{Shopping Interest}: Summarize the product information.

\item\textbf{Brand Preference}: Choose from the product information, and only keep brand names.

\end{itemize}

Second, summarize your personal preferences across the following aspects:

\setlength{\leftmargini}{10pt} \begin{itemize}
\item\textbf{Diversity Preference}: 

Do you prefer trying new things or sticking to familiar products? 
Choose from <DIVERSITY>.

\item\textbf{Interaction Complexity}

Do you prefer simple and concise interactions, or do you enjoy detailed and thorough exchanges? 
Choose from <INTERACTION>.

\item \textbf{Tone and Style}

Summarize your overall emotional tone, speaking style, and expressive characteristics when giving reviews. Keep keywords.

\item \textbf{Item Reference}

Summarize your tendency to refer to specific products or brands in your reviews, like purchase history, shopping cart, or recommendations from friends. Keep keywords.

\item\textbf{Focus Aspect}

What aspects of products do you pay more attention to? 
Choose from <FOCUS ASPECT> or summarize others from reviews. Keep keywords.

\end{itemize}

In the end, arrange all the above aspects using the JSON format, with each aspect as an individual key.

Do not include any additional information or explanations and stay grounded.

Your History:

<HISTORY>

}

\caption{User profile generation.}
\label{app: prompt profile generation}
\end{figure*}
\begin{figure*}
\FancyBox{Details of Choices in Profile Generation}{
\setlength{\leftmargini}{10pt} \begin{itemize}
\item \textbf{<Gender>:} 

[Female, Male].
\item \textbf{<AGE>:} 

[Under 18, 18-24, 25-34 , 35-44 , 45-49 , 50-55 ,  56+].
\item \textbf{<OCCUPATION>:} 

[Academic/Educator, Artist, Clerical/admin, College/grad student, Customer service, Doctor/health care, Executive/managerial, Farmer, Homemaker, K-12 student, Lawyer, Programmer, Retired, Sales/Marketing, Scientist, Self-employed, Technician/Engineer, Tradesman/Craftsman, Unemployed, Writer, Other].
\item \textbf{<PRICE SENSITIVITY>:} 

["High": "A Price-Conscious Shopper who is very sensitive to cost and seeks the best deals.",

"Medium": "A Balanced Buyer who considers price but also values quality and features.",

"Low": "A Value-Driven Consumer who prioritizes quality and features over price."].
\item \textbf{<DIVERSITY>:}

[
"High": "A Highly Adventurous Explorer eager to discover diverse products across categories. They often seek recommendations, and purchase a wide variety of items with varying ratings, and the user's own ratings may often differ from the average. Their reviews are detailed and enthusiastic, reflecting their unique tastes and enjoyment of variety.",

"Medium": "A Balanced Seeker who enjoys trying new products but also values familiarity. They appreciate targeted recommendations, purchase a moderate number of items with solid ratings and a reasonable number of ratings, and their reviews balance detailed feedback with concise, practical comments.",

"Low": "A Meticulously Selective Buyer who sticks to tried-and-true products, showing little interest in new options. They purchase fewer items, favoring those with high ratings and a large number of ratings. Their own ratings are often very close to or slightly above the average, and their reviews are thoughtful and focused on familiar products."
].
\item \textbf{<INTERACTION>:}

[
"High": "A Thorough Conversationalist who enjoys detailed discussions, exploring all aspects of a product or service. They provide extensive reviews and value comprehensive support, engaging in multiple rounds of communication.",

"Medium": "A Moderate Engager who balances simplicity with detail. They prefer clear communication but can engage in detailed exchanges when necessary. They provide reviews that are a mix of concise observations and some detailed insights, especially if they have strong feelings about a product.",

"Low": "A Minimalist Interactor who values simplicity and efficiency. They prefer quick, straightforward interactions and leave brief, to-the-point reviews, focusing only on essential product aspects."
].
\item \textbf{<FOCUS ASPECT>:}

["Average Rating", "Number of Ratings", "Price", "Store", "Material", "Size", "Weight", "Brand"].
\end{itemize}
}
\caption{Details of choices in user profile generation.}
\label{app: prompt profile generation cho}
\end{figure*}

\begin{figure*}
\FancyBox{Search Instruction Generation}{
You will act as an online shopper.

Your Profile:

<PROFILE>

You are looking for a product similar to the following product:

<PRODUCT>

You want to find a similar product, but you are not looking for an exact match.

Generate a search request that is somewhat vague, reflecting your preferences and personalities without revealing the complete details of the target product.

Rules:
\setlength{\leftmargini}{10pt} \begin{itemize}
\item <DIVERSITY>.
\item <INTERACTION>.
\item You pay more attention to <FOCUS\_ASPECT> of products, make sure to include some of them in the search request.
\item Ensure the search request aligns with your overall tone and style: <TONE\_AND\_STYLE>.
\item Do not repeat the exact information in your profile or product. Instead, use your own words to convey the same information.
\item Try to make the request as natural as possible and stick to the personalities in your profile.
\item Do not include any additional information or explanations and stay grounded.
\item Do not hallucinate information that is not provided,
\item No more than <NUM> words.
\end{itemize}

}
\caption{Search instruction generation.}
\label{app: prompt search generation}
\end{figure*}
\begin{figure*}
\FancyBox{Recommendation Instruction Generation}{
You will act as an online shopper.

Your Profile:

<PROFILE>

You have recently shown interest in the following type of product: 

<PRODUCT>

Now, you're exploring options that could match your overall tastes, but you're not sure exactly what you're looking for. 

Generate a recommendation request that reflects your general preferences and style, but leaves room for flexibility and discovery.

Rules:
\setlength{\leftmargini}{10pt} \begin{itemize}
\item <DIVERSITY>.
\item <INTERACTION>.
\item You value <FOCUS\_ASPECT> of products, but keep the request open-ended to allow for a variety of recommendations.
\item Ensure the recommendation request aligns with your overall tone and style: <TONE\_AND\_STYLE>.
\item Do not restate your profile or the product. Use different words or hints to convey your preferences.
\item Avoid being too specific or precise in your request.
\item Try to make the request as natural as possible and stick to the personalities in your profile.
\item Do not include any additional information or explanations and stay grounded.
\item Do not hallucinate information that is not provided,
\item No more than <NUM> words.
\end{itemize}

}
\caption{Recommendation instruction generation.}
\label{app: prompt rec generation}
\end{figure*}
\begin{figure*}
\FancyBox{Review Instruction Generation}{
You will act as an online shopper.

Your Profile:

<PROFILE>,

You have recently purchased the following product: 

<PRODUCT>,

Your feelings about the product are:

<REVIEW>,

Now, You want to write a review.

Generate a review request to ask for assistance to create a complete review that reflects your preferences and typical review style.

Rules:
\setlength{\leftmargini}{10pt} \begin{itemize}
\item <INTERACTION>.
\item You value <FOCUS\_ASPECT> of products, but keep the request open-ended to allow for a variety of recommendations.
\item Ensure the review request aligns with your overall tone and style: <TONE\_AND\_STYLE>.
\item Do not simply restate your profile or feedback verbatim. Instead, paraphrase and expand to reflect a more comprehensive review.
\item Try to make the request as natural as possible and stick to the personalities in your profile.
\item Do not include any additional information or explanations and stay grounded.
\item Do not hallucinate information that is not provided.
\item No more than <NUM> words.
\end{itemize}

}
\caption{Review instruction generation.}
\label{app: prompt review generation}
\end{figure*}

\begin{figure*}
\FancyBox{User Simulation Instruction}{
You are a user interacting with a personalized shopping agent.

Your Profile:

<PROFILE>

You have purchased the following product:

<PRODUCT>

(In review tasks:)
Your review is as follows:

<REVIEW>

The shopping agent will help you complete your shopping requests.

Rules:
\setlength{\leftmargini}{10pt} \begin{itemize}
\item Just generate one line at a time to simulate the user's message.
\item Do not hallucinate information that is not provided.
\item Do not give additional instructions or ask questions, only respond to the agent's questions.
\item Do not provide any specific product details.
\item Do not repeat the exact information in your profile or product. Instead, use your own words to convey the same information.
\item Try to make the conversation as natural as possible and stick to the personalities in your profile.
\item If the result is not satisfactory, you can express your dissatisfaction and provide clues to help the agent understand your preferences.
\end{itemize}

}
\caption{User simulation instruction.}
\label{app: prompt user simulation}
\end{figure*}

\begin{figure*}
\FancyBox{Prompt in Single-Turn Track}{
As a personalized shopping agent, you can help users search for products, recommend products, or complete their reviews.

Rules:

\setlength{\leftmargini}{10pt} \begin{itemize}
\item The user will provide user\_id and a request.
\item You need to use the most appropriate tool to find the product or fill the review that matches the user's request.
\item You are not allowed to interact with the user. Make the best tool call based on the user's request.
\item You have only one chance to make a tool call, so make sure you have the best input for the tool.
\item The tool will be provided, you need to use the tool and provide the most appropriate input for the tool. Do not use other tools.
\item Formulate the best input for the tool based on the user's request and the memory provided.
\end{itemize}

Memory:

(Depends on the  method.)

Functions to use:

(Available functions and their descriptions.)

}
\caption{Prompt template for all methods in single-turn track.}
\label{app: single prompt}
\end{figure*}

\begin{figure*}
\FancyBox{Prompt in Multi-Turn Track}{
As a personalized shopping agent, you can help users search for products, recommend products, or complete their reviews.

Rules:

\setlength{\leftmargini}{10pt} \begin{itemize}
\item The user will provide user\_id and a request.
\item You need to use the tools to find the product or fill the review that matches the user's request.
\item The tool will be provided, you need to use the tool and provide the most appropriate input for the tool. Do not use any other tools.
\item Formulate the best input for the tool based on the user's request and the memory provided.
\item You are allowed to interact with the user by 'respond' to ask for more information or feedback, but steps are limited, and less steps are preferred.
\item Your main goal is to help the user complete the task as accurately and efficiently as possible, do not keep responding to the user, focus on making the better tool calls.
\item The evaluation will be based on the ranking of the target product in search and recommendation tasks, and the similarity of the review in the review task.
\item When you think you have found the best input for the task tool calls, you can end the task by making a 'stop' call. 
\item You should not make up any information or knowledge not provided from the user or the tools, or give subjective comments or recommendations.
\item You should at most make one tool call at a time, and if you take a tool call, you should not respond to the user at the same time. If you respond to the user, you should not make a tool call.
\end{itemize}

Memory:

(Depends on the  method.)

Functions to use:

(Available functions and their descriptions.)

}
\caption{Prompt template for all methods in multi-turn track.}
\label{app: multi prompt}
\end{figure*}

\end{document}